\documentclass[conference]{IEEEtran}
\IEEEoverridecommandlockouts
\usepackage[T1]{fontenc}
\usepackage[utf8x]{inputenc}
\usepackage{cite}
\usepackage{amsmath,amssymb,amsfonts}
\usepackage{algorithmic}
\usepackage{graphicx}
\usepackage{textcomp}
\usepackage{url}
\usepackage{xcolor}
\usepackage{algorithm}
\usepackage{algorithmic}
\def\BibTeX{{\rm B\kern-.05em{\sc i\kern-.025em b}\kern-.08em
    T\kern-.1667em\lower.7ex\hbox{E}\kern-.125emX}}
\begin{document}

\title{A Privacy-Preserving Unsupervised Domain Adaptation Framework for Clinical Text Analysis}


\author{Qiyuan An,
        Ruijiang Li,
        Lin Gu,
        Hao Zhang,
        Qingyu Chen,
        Zhiyong Lu,
        Fei Wang,
        and~Yingying Zhu 
\IEEEcompsocitemizethanks{\IEEEcompsocthanksitem Q. An and Y. Zhu are with the Department of Computer Science and Engineering, University of Texas at Arlington. Email: {qxa5560@mavs.uta.edu, yingying.zhu@uta.edu}
\IEEEcompsocthanksitem R. Li is with eBay Inc. Email: {ruijili@ebay.com}
\IEEEcompsocthanksitem L. Gu is with RIKEN, AIP / The University of Tokyo. Email: {lin.gu@riken.jp}
\IEEEcompsocthanksitem H. Zhang and F. Wang are with Cornell University. Email: {haz4007@med.cornell.edu, few2001@med.cornell.edu}
\IEEEcompsocthanksitem Q. Chen and Z. Lu are with NIH. Email: {qingyu.chen@nih.gov, zhiyong.lu@nih.gov}
}
}

\maketitle

\begin{abstract}
Unsupervised domain adaptation (UDA) generally aligns the unlabeled target domain data to the distribution of the source domain to mitigate the distribution shift problem. The standard UDA requires sharing the source data with the target, having potential data privacy leaking risks. To protect the source data's privacy, we first propose to share the source feature distribution instead of the source data. However, sharing only the source feature distribution may still suffer from the membership inference attack who can infer an individual's membership by the black-box access to the source model. To resolve this privacy issue, we further study the under-explored problem of privacy-preserving domain adaptation and propose a method with a novel differential privacy training strategy to protect the source data privacy. We model the source feature distribution by Gaussian Mixture Models (GMMs) under the differential privacy setting and send it to the target client for adaptation. The target client resamples differentially private source features from GMMs and adapts on target data with several state-of-art UDA backbones. With our proposed method, the source data provider could avoid leaking source data privacy during domain adaptation as well as reserve the utility. To evaluate our proposed method's utility and privacy loss, we apply our model on a medical report disease label classification task using two noisy challenging clinical text datasets. The results show that our proposed method can preserve source data's privacy with a minor performance influence on the text classification task.
\end{abstract}


\section{Introduction}
\begin{figure}[t]
\centering
\includegraphics[width=0.5\textwidth]{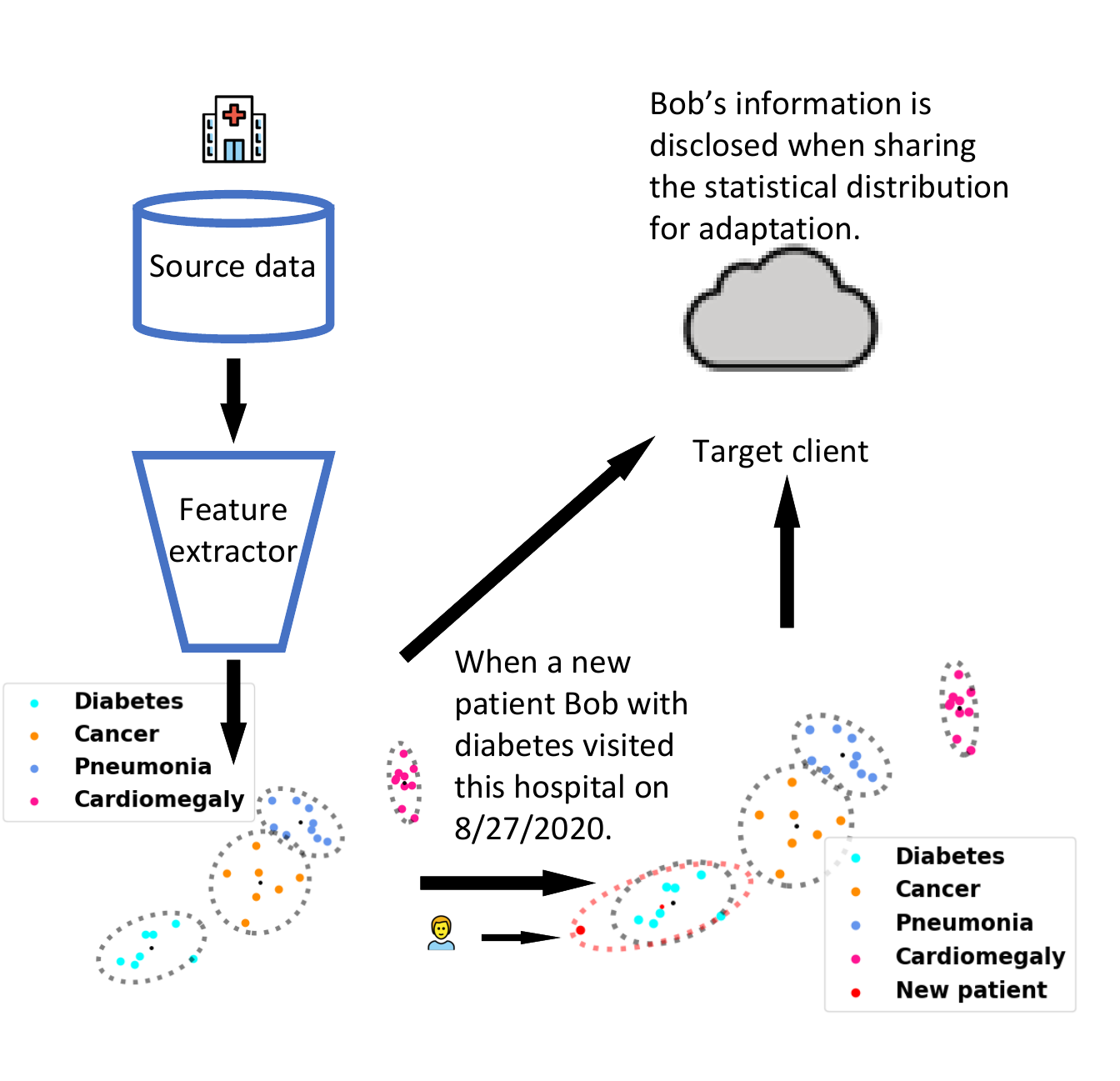} 
\caption{The membership inference attack scenario when sharing the source feature distributions to the target client. } 
\label{fig:attack}
\end{figure}
\par
Unsupervised domain adaptation (UDA) has been explored broadly to address the data domain-wise distribution shift and transfer the source model to unlabelled target with impressive results in a wide variety of areas, including computer vision \cite{lee2019sliced,saito2018maximum,kang2019contrastive,DAmedical}, natural language processing (NLP) \cite{tzeng2017adversarial,dai2020adversarial,peddinti2011domain}, and others \cite{ghosal2020kingdom,ma2019domain}. The general idea of unsupervised domain adaptation is to learn a domain-invariant feature space by aligning the target data's feature distribution to the source data's feature distribution. Several different feature alignment approaches were developed including adversarial domain adaptation \cite{ghosal2020kingdom}, maximum mean discrepancy \cite{lee2019sliced,saito2018maximum}, contrastive learning guided feature alignment \cite{kang2019contrastive}, and others \cite{liang2020we}.
\par
While there are numerous publications about the UDA in computer vision tasks \cite{DAmedical,liang2019distant,long2018conditional,ganin2015unsupervised,kang2019contrastive,saito2018maximum,lee2019sliced}, the UDA in text analysis remains an under-explored area. Directly applying the UDA methods on the text data has several difficulties: 1) the text data are discrete so that domains in text data are more ambiguous \cite{du-etal-2020-adversarial}; 2) pretrained text encoder such as BERT \cite{devlin2018bert} is prone to overfitting among domains due to the large model size \cite{BERTAAD}. Consequently, designing a UDA method to generalize on the text data requires meticulous attention.
\par
Besides the general problems in text UDA, the data privacy concern is another big issue \cite{privacy}. In the general UDA setting, the pretrained source model and source data are shared to target clients for training a domain adaptation model. However, this setting not only raises the data privacy concern on source data but sharing the source data, e.g., from some medical institutions, is not even allowed by laws or regulations \cite{jochems2017developing,jochems2016distributed}. Previously, only a few works have explored the source privacy-preserved method to protect the privacy of source data by sharing the source feature's distribution instead of the source data \cite{liang2020we,yeh2021sofa,DAmedical}. However, 
these methods are still vulnerable to the membership inference attack \cite{osuolale2017secure,odun2017overview,shokri2017membership}, allowing the attackers to reveal an individual's membership in the source dataset from the disclosure of overall statistical information \cite{dwork2008differential,shokri2017membership,rahman2018membership}.

\par
Figure~\ref{fig:attack} shows an example of the membership inference attack under the UDA setting when sharing the source feature's distribution to target clients. A hospital database maintains the populations of many different diseases and only provides some features with desensitized information to the outside, e.g., the distribution of each particular disease. Intuitively, the released distribution won't compromise each individual's privacy.
However, an adversary could reveal an individual's physical condition by comparing disease distributions before and after that patient's visit, i.e., if the diabetes population grows by one or more generally the feature distribution of the diabetes changes after that person's visit, we can infer this new patient has diabetes. 
\par

To deal with the issues described above, we propose a privacy-preserving UDA method. At the algorithm level, we share the high-level source feature's distribution modeled by Gaussian Mixture Models (GMMs) instead of the source data along with the source feature extractor and source classifier to the target client. The target client will resample new source features from the GMM and train the domain adaptation model locally. To further defend against the membership inference attack on the statistic information of source data (source feature's GMM), we further propose to pretrain the source feature extractor with differential privacy (DP) as DP is a \cite{dwork2006calibrating} gold standard of protecting the training data's privacy against membership inference attack. In the end, the DP-pretrained source feature extractor with GMM can be shared with the target client to facilitate any form of domain adaptions.
\par
Our proposed method, the Privacy-Preserving Unsupervised Domain Adaptation Framework with Differential Privacy is shown in Figure \ref{fig:frame}, has the following contributions:

\begin{itemize}

\item 
To resolve the BERT overfitting issue as mentioned above, we opt for the knowledge distillation (KD) loss by smoothing the features of target data between source and target models \cite{BERTAAD}.

\item To protect the source data privacy in UDA, we propose to model source feature's distributions in GMMs and use the resampled source feature in target client for domain adaptation training. To further protect from leaking the individual information from the shared source feature distribution, we design the DP-pretrained source BERT model following \cite{abadi2016deep}.
By sharing only the source feature distribution, we (substantially) limit the privacy loss of source data within a predefined privacy budget. Note that DP cannot guarantee the data privacy absolutely but can ensure only a limited amount of risk is exposed by sharing the source feature distribution in our case \cite{dwork2008differential}.

\item Our proposed method is compatible with many UDA methods. In our implementation, we adopt two state-of-art UDA backbones: Domain-Adversarial Training of Neural Networks (DANN) \cite{ganin2015unsupervised}, and Conditional Adversarial Domain Adaptation (CDAN) \cite{long2018conditional}.

\item We firstly study a very challenging task: transferring a disease label extraction model to unlabelled clinical text, which has great potential to be applied to improve real-world clinical workflow. Many clinical text datasets are created by rule-based automatic labeling tools to save time and reduce cost as labeling the clinical text is extremely expensive due to requiring expert knowledge and excess labor \cite{johnson2019mimic}. However, these auto-labeled clinical text datasets contain a large number of noisy labels for clinical text \cite{hao2020inaccurate}.

\item We evaluate our approach on two clinical text disease-label extraction datasets. Our method even outperforms the source data available method significantly, suggesting our method is robust to noisy labels.

\end{itemize}

\section{Related Work}

\subsection{Unsupervised Domain Adaptation with Sharing Source Data}
UDA aims to transfer a source model (the feature extractor and the classifier) to the target domain with only unlabelled data by learning a shared domain invariant feature space to resolve the feature distribution shift problem among different domains \cite{long2015learning,long2016unsupervised,tzeng2014deep}. It is generally achieved by minimizing the distribution discrepancy between source and target data with the following methods: maximum mean discrepancy (MMD) \cite{long2015learning,long2016unsupervised,tzeng2014deep,long2018conditional,kang2019contrastive}, matching the distribution statistical moments at different order \cite{sun2016deep,chen2020homm} , minimum global transportation cost \cite{bhushan2018deepjdot,xu2020reliable}, adversarial learning \cite{ganin2015unsupervised} and conditional adversarial learning \cite{long2018conditional,pan2019transferrable,gu2020spherical}. However, all these domain adaption methods require sharing the source data to target clients, which are not suitable for privacy-sensitive data. Moreover, sharing source and target data online could lead to potential security issues, and expensive network transmission costs \cite{privacy}.

\subsection{Source Data-Protected Domain Adaptation}
Source data-protected domain adaption has been proposed to protect source data privacy with only the source classifier and feature extractor provided to the target client for domain adaption, including two major types: 1) Two-stage shared source and target feature extractor method. It first clusters the source features by class labels, then uses unsupervised clustering methods to align target features to the source feature clusters \cite{chidlovskii2016domain,liang2019distant}. However, shared source and target feature extractors are only suitable for the small domain distribution shift problem. 2) Separate source and target feature extractors. \cite{tzeng2017adversarial} firstly proposed to train separate feature extractors on the source and target domain for the significant domain shift problem. Several end-to-end source data-protected domain adaption methods were developed by learning the target domain feature extractor that automatically clusters target features (unsupervised or self-supervised) and maximizes the information maximization (IM) loss on the source classifier to find the optimal target decision boundary \cite{liang2020we,li2020model}. Unfortunately, these source data-protected DA methods are still vulnerable to the membership inference attack who can infer the individual's information from the shared source model and classifier.

\subsection{Domain Adaptation with Privacy Concern}
Some previous literature has studied the data privacy issue under the DA setting. \cite{song2020privacy} protects the data privacy without sharing the sensitive data between source and target domains by aggregating small source models into a central server in the federated setting. \cite{li2020multi} proposed randomizing the shared local model weights to protect from gradient leakage attack. \cite{wang2020deep} performed domain adaptation in an adversarial-training manner. However, \cite{lyu2020differentially} has shown that although adversarial training can make it difficult for the adversary to recover the input data, it cannot entirely remove the sensitive information from the data representations. To the best of our knowledge, leveraging differential privacy to protect source data's privacy has not been studied so far. We firstly study the DP-based source privacy-preserved UDA method in this work.
\par
Differential privacy (DP) provides a strong privacy guarantees for algorithms on aggregate databases \cite{dwork2010boosting,dwork2014algorithmic,abadi2016deep}. In our application, the source dataset consists of many text-label pairs. According to the DP definition, two datasets are adjacent if they differ in one element, i.e., one text-label pair is added or removed from the source dataset.
\par
\textit{Definition 1.} A randomized mechanism $\mathcal{M}: \mathcal{D}\rightarrow \mathcal{R}$ maps an entry of $D$ to $\mathcal{R}$. The mechanism $\mathcal{M}$ is $(\epsilon,\delta)$-differential privacy if for any two adjacent inputs $d, d'\in \mathcal{D}$ and any outputs $S\in \mathcal{R}$ it holds that 
\begin{equation}
\begin{aligned}
Pr[\mathcal{M}(d)\in S]\leq e^{\epsilon}Pr[\mathcal{M}(d')\in S]+\delta.
\end{aligned}
\end{equation}
From the definition by Dwork et al. \cite{dwork2006calibrating}, the term $\epsilon$ quantifies how much we tolerate the output of a model on one database $d$ to be different from the outputs of the same model on an adjacent database $d'$ that only has one different training sample, while $\delta$ is the failure rate which we tolerate the guarantee to not hold. 
\par
Another property of DP is particularly useful for our application is the post-processing property that any computation applied to the output of an $(\epsilon,\delta)$-DP algorithm maintains $(\epsilon,\delta)$-DP \cite{dwork2014algorithmic,lyu2020differentially}. This property allows us to apply any post-processing algorithms to the DP-pretrained source model without sacrificing its DP property.

\subsection{Cross-Domain Text Classification}
Most of the above DA works focused on image classification tasks with models pretrained on Imagenet. The text-based domain adaptation is less explored and has several differences compared to image domain adaptation. Firstly, different imaging domains are easily separable (such as natural images and sketch images), and therefore the feature extractor has less chance to overfit between two domains. The domains of text data are ambiguous since similar text might have different meanings on different domains~\cite{du-etal-2020-adversarial}\cite{ghosal2020kingdom}. To this end, a pretrained deep learning text analysis model, such as BERT, is likely to overfit target and source domains. Yu et al. \cite{BERTAAD} used the knowledge distillation (KD) loss to prevent the overfitting of BERT on the target data with significant performance improvement. We follow a similar framework using the pretrained BERT and KD loss to address the overfitting issue.

\section{Method}
This section presents the building blocks of our proposed method. We begin by introducing the general UDA framework, then the source GMM resampling, followed by pretraining source feature extractor with DP, next the KD loss to smooth the BERT outputs, and a small trick of information maximization to enhance UDA. In the end, we wrap up the privacy-preserving UDA with the DP framework.
\subsection{Unsupervised Domain Adaptation Notations}
Given a source dataset $\mathcal{X}_s=\{\mathbf{x}_i^s\}_{i=1}^{n_s}$ sampled from the source domain $\mathcal{D}_s$ with labels $\mathcal{Y}_s =\{\mathbf{y}^s_i\}_{i=1}^{n_s}, \mathbf{y}_s^i \in \{1, 2, ..., K\}$ for $K$ different classes and an unlabeled target dataset $\mathcal{X}_t=\{\mathbf{x}_i^t\}_{i=1}^{n_t}$ sampled from the target domain $\mathcal{D}_t$. The source domain $\mathcal{D}_s$'s distribution is different from the target domain $\mathcal{D}_t$'s distribution. Most current DA methods assume there is a shared domain invariant feature space $\mathcal{Z}$ and they can learn feature extractors $E_s:\mathcal{X}_s\to \mathcal{Z}$ and $E_t:\mathcal{X}_t \to \mathcal{Z}$ mapping the source and target data into the shared space ~$\mathcal{Z}$.
The same classifier $C:\mathcal{Z} \to \mathcal{Y}$ trained on the labeled source data can be applied to the target data in the shared feature space $\mathcal{Z}$. Both $E$ and $C$ are neural networks parameterized by $\theta_{E}$  and $\theta_{C}$.
\par
The classic domain adaptation methods can be divided into two categories: 1) to learn a shared feature extractor on both source and target domains, i.e., $E_t = E_s=E$, which only works well when the domain shift is negligible \cite{chidlovskii2016domain, liang2019distant, ganin2015unsupervised}. 2) to learn separate feature extractors $E_s$ (source) and $E_t$ (target) on different domains, which has shown to successfully outperform the previous kind of domain adaptation methods \cite{tzeng2017adversarial} on image classification tasks.
The general loss function of the second kind can be written as follows with two parts:
\begin{eqnarray}
\nonumber\mathcal{L}_{DA} (\theta_{E_s},\theta_{E_t},\theta_{C_s}) = \mathcal{L}_{cls}(C_s(E_s(\mathcal{X}_s)), \mathcal{Y}_s) \\
+ \mathcal{L}_d(E_s(\mathcal{X}_s), E_t(\mathcal{X}_t)),
\label{eq:da_loss}
\end{eqnarray}
where the first term is to minimize the classification error on source data with labels by the cross-entropy loss. When $E_t=E_s=E$, it will learn a shared feature extractor for both source and target domains. The second term aims to minimize the feature distribution difference between source and target data through different metrics such as  KL-divergence \cite{tzeng2017adversarial}, MMD distance \cite{ganin2015unsupervised}, and Wasserstein-distance \cite{xu2020reliable, bhushan2018deepjdot}. The above loss function Eq. \ref{eq:da_loss} has been successfully applied to plenty of UDA problems when both source and target data are accessible. In practice, the source feature extractor $E_s$ is always used to initialize the target feature extractor~$E_t$. 
\begin{figure}[t]
\centering
 \includegraphics[width=0.5\textwidth]{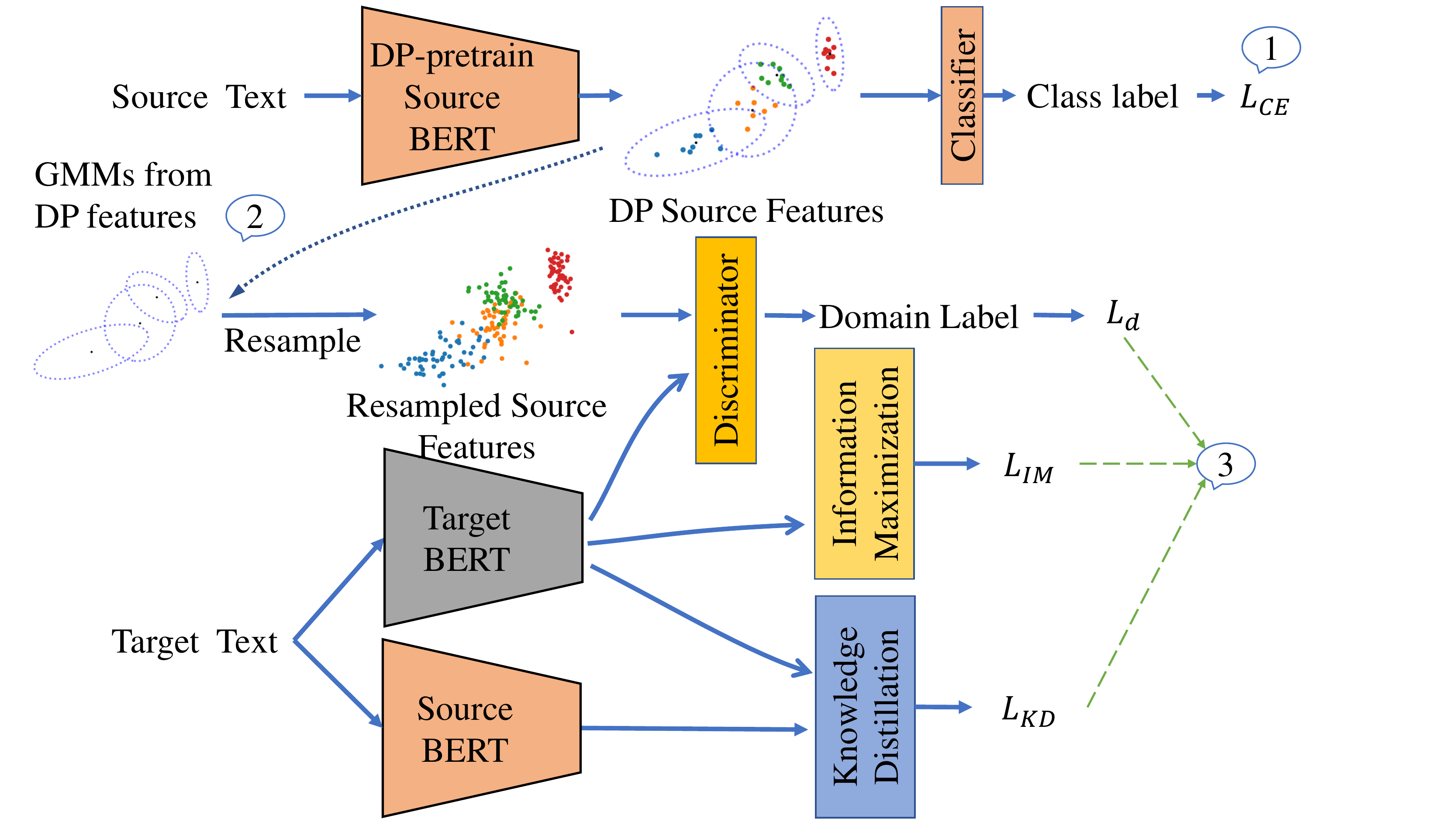}
\caption{Our proposed Differentially Private Source Privacy-Preserving Domain Adaptation Framework. Step 1: Pretrain the source BERT and classifier with DP. Step 2: Extract GMMs from DP source features. Step 3: Adaptation with domain loss, information maximization loss, and knowledge distillation loss.} 
\label{fig:frame}
\end{figure}

\subsection{Source Data Protected Domain Adaptation}
Instead of sharing the source data directly to the target client for adapting, we send the source feature's distribution to the target client and then the target client resamples features from source feature distributions to adapt the target feature extractor $E_t$. Let $p_s(z)$ define the source feature's distribution, where $p_s(z)$ can be parameterized by different types of probabilistic density function. Empirically, we find that the high-level features of the source domain are nicely clustered into different classes after pretraining. Hence, we apply the GMM as the probabilistic density function in our experiments. Here, the GMM can be replaced by various other probabilistic density functions instead.
Our proposed source data-protected UDA loss function becomes:
\begin{equation}
\begin{aligned}
 {\mathcal{L}_d} (\theta_{E_t},\theta_{p_s}) &=\mathcal{L}_{cls}(C_s(E_s(\mathcal{X}_s)), \mathcal{Y}_s) \\ 	
&+ {\mathcal{L}_d}_{z_i^s\sim p_s(z) }(\mathcal{Z}_s, E_t(\mathcal{X}_t)),
	\label{eq:da_loss2}
 \end{aligned}
\end{equation}
where $p_s(z)$ represents the high-level source feature distribution learned by BERT on source data (we use GMM to model $p_s(z)$, the parameters of $p_s$ are denoted as $\theta_{p_s}$).  $\theta_{p_s} = [\mathbf{\mu}^s_k,\mathbf{\Sigma}_k^s \in \mathcal{R}^{d_s\times K_s}, \mathbf{\pi}_k^s\in \mathcal{R}^{K_s\times 1} ]$ are the source distribution's means, co-variances, and the Gaussian mixture parameters respectively.
$\mathcal{Z}_s =\{\mathbf{z}_i^s\}_{i=1}^{N_s}$ is the set of resampled source features with $N_s$ samples, $\mathbf{z}_i\in\mathcal{R}^{d_s\times 1}$. 
$d_s$ is the source feature dimension, $K_s$ is the number of classes in the source domain. Source model $E_s, C_s$ are pretrained on source data and fixed. Then we only optimize the target feature extractor $E_t$ during adapting.

\subsection{DP-Pretrained Source Protected DA}
Although sharing only the source feature's GMM to the target client does not reveal exact data representations, an adversary could still reveal the membership of a sample by comparing the changes from source feature distributions through the membership inference attack as shown in Figure \ref{fig:attack}. To further protect the source data's privacy, we opt for the differential-privacy training process on pretraining the source feature extractor. Denote the DP-pretrained source feature extractor as $E_s^{dp}$ and the source feature's GMM as $p_s(z)^{dp}$, the objective function of adapting the target feature extractor then becomes:
\begin{equation}
\begin{aligned}
 {\mathcal{L}_d} (\theta_{E_t},\theta_{p_s}) &=\mathcal{L}_{cls}(C_s(E^{dp}_s(\mathcal{X}_s)), \mathcal{Y}_s) \\ 	
&+ {\mathcal{L}_d}_{z_i^s\sim p_s(z)^{dp} }(\mathcal{Z}_s, E_t(\mathcal{X}_t)).
	\label{eq:da_loss3}
 \end{aligned}
\end{equation}

Since our goal is to protect the source data's privacy while maintaining the pretrained model's utility, we adopt the DP-SGD method \cite{abadi2016deep} and Rényi Differential Privacy (RDP) \cite{mironov2017renyi,mironov2019r} for the differential privacy-protected pretraining process. The DP-SGD method tracks the privacy loss by moments accountant (MA) to obtain a tighter estimate of the overall privacy loss compared to the strong composition theorem \cite{dwork2010boosting}. 

\subsection{Privacy Analysis of DP-Pretraining Source Feature Extractor}
\begin{algorithm}[h]
\caption{DP Pretraining the Source Feature Extractor}
\label{alg:dp_pretrain}
\textbf{Input}: Source samples $\{x_1^s, ..., x_N^s\}$, loss function $\mathcal{L}(\theta_{E_s})=\frac{1}{N}\sum_i\mathcal{L}(\theta_{E_s}, x_i^s)$.\\
\textbf{Hyperparameters}: 
number of accumulated steps $L$, total number of iterations $T$.\\
\textbf{Output}: $\theta_{E_s}$ and accumulated privacy loss in terms of ($\epsilon,\delta$).
\begin{algorithmic}[1] 
\STATE Initialize $\theta_{E_s}$ as a pretrained BERT-base-cased text classification model
\WHILE{t $<$ T}
\IF{l $<$ L}
\STATE Randomly sample batch size samples $B_t$.
\STATE Compute per example gradient with clip by max norm and add noise: \\ $g_l(x_i^s)\leftarrow\nabla_{\theta_{E_s}}\mathcal{L}(\theta_{E_s}, x_i^s)$ with $x_i^s\in B_t$.

\STATE Accumulate gradients without updating: \\
$g_t\leftarrow \sum_l g_l(x_i^s)$.
\ELSE
\STATE Perform gradient updating: \\
$\theta_{E_s,t+1}\leftarrow AdamW(\theta_{E_s,t},g_t)$
\ENDIF
\STATE Compute privacy spent on each iteration by the MA method.
\ENDWHILE
\STATE \textbf{return} $\theta_{E_s,T}$ and the overall privacy loss ($\epsilon,\delta$).
\end{algorithmic}
\end{algorithm}

Algorithm \ref{alg:dp_pretrain} outlines our proposed pretraining method to protect the source data privacy as well as provide an accurate accountant of privacy loss. We opt for the DP-SGD algorithm \cite{abadi2016deep} to pretrain the source feature extractor with parameters $\theta_{E_s}$ on the clinical text classification task by the cross-entropy loss function $\mathcal{L}(\theta_{E_s})$ to produce a source model that satisfies $(\epsilon,\delta)$-differential privacy. At each step of the backpropagation, we derive the gradient $\nabla_{\theta_{E_s}}\mathcal{L}(\theta_{E_s}, x_i^s)$ for each sample in a randomly sampled mini-batch, clip the $l_2$ norm of each gradient, add noise to each gradient and compute the average, accumulate a few steps without updating the parameters $\theta_{E_s}$, and finally update the parameters $\theta_{E_s}$ towards reducing the cross-entropy loss $\mathcal{L}(\theta_{E_s})$. After each step, we also need to compute the privacy loss based on the moments accountant (MA) method \cite{abadi2016deep} to obtain the final accumulated privacy loss in terms of ($\epsilon,\delta$).
\par
According to the DP-SGD's theorem 1 \cite{abadi2016deep}, the per-example gradient $g_l(x_i^s)$ is added by the Gaussian noise of $\mathcal{N}(0,\sigma^{2}C^{2}I)$. There exists constants $c_{1}$ and $c_2$ so that given the sampling probability $q=L/N$ and the number of steps $T$, for any $\epsilon<c_{1}q^{2}T$, our proposed algorithm \ref{alg:dp_pretrain} is also $(\epsilon,\delta)$-DP for any $\delta>0$ if we choose 
\begin{equation}
\begin{aligned}
\sigma\geq c_{2}\frac{q\sqrt{T\log(1/\delta)}}{\epsilon}. \nonumber
\end{aligned}
\end{equation}
From the above theorem proved in \cite{abadi2016deep}, we can achieve $(\epsilon,\delta)$-DP in protecting the source data's privacy when using the MA to track the privacy loss. From the post-processing property of DP-pretraining \cite{dwork2014algorithmic,lyu2020differentially}, the following GMM resampling step will maintain the $(\epsilon,\delta)$-DP property.

\subsection{Smoothing the Source and Target BERT Outputs by Knowledge Distillation}
The issue in Eq. \ref{eq:da_loss2} is that the learned feature extractor $E_t$ may overfit the source data and perform poorly on the target data. Yu et al. \cite{BERTAAD} has shown that adopting the KD loss into the domain adaptation framework with separate feature extractors can mitigate this overfitting issue of the target encoder, and their experiments further confirm this point. However, their original design smoothed the outputs of $E_s, E_t$ through the source data, where the source data is unavailable in our setting. Therefore, we modify their method by replacing the source data with the target data to adopt the KD loss on the target encoder. Note that this is different from the original version of KD \cite{hinton2015distilling} who learns a smaller model using the soft labels inferred from a teacher model. We only utilize the KD loss function to smooth outputs from the target encoder. Thus, our objective function of KD loss becomes this: 
\begin{equation}
\begin{aligned}
\mathcal{L}_{KD}(x_{j}^{t}) &= t^{2} \mathbb{E}_{x_{t}\sim
\mathcal{X}_{t}}\sum_{j=1}^{N_t}-softmax(\frac{C(E_{s}(x_{j}^{t}))}{t})  \\
&\cdot\log(softmax(\frac{C(E_{t}(x_{j}^{t}))}{t})),
\end{aligned}    
\end{equation}
where $t$ is the temperature hyperparameter measuring the smoothness of the output labels using different feature extractor $E_s, E_t$, $\mathbb{E}$ represents taking the average of target samples. The temperature $t$ is chosen by the experiments from \cite{BERTAAD} that $t=20$ can balance between maintaining the class feature and smoothing the distributions of class outputs. Our experiment results have validated that this swap has the same effect as using the source data for KD loss.
\par

Since the benefit of adopting the KD loss could enhance the target encoder's generalization, we also modify two other domain adaptation methods, DANN and CDAN, by integrating the KD loss into their adapting process. From our experiments, the KD loss can significantly improve the DANN method's performance on text data. Details can be found in the next section.

\subsection{Information Maximization on Target Data}
In the UDA setting, the target data are unlabeled, and therefore we cannot apply the supervised classification signal to train the target feature extractor. To overcome this difficulty, we also adopt the information maximization (IM) loss as Eq. \ref{eq:im_loss} $\mathcal{L}_{IM}=\mathcal{L}_{ent}+\mathcal{L}_{div}$ on the target features from \cite{gomes2010discriminative} to make the target intra-class features converge and inter-class features diverge. 
\begin{align}
\mathcal{L}_{ent}&=-\mathbb{E}_{x_t\sim \mathcal{X}_t}\sum_{i=1}^{K}(C(E_{t}(x_{t})))\log(C(E_{t}(x_{t}))), \nonumber \\
\mathcal{L}_{div}&=\sum_{i=1}^{K}g_{k}\log(g_{k})
\label{eq:im_loss}
\end{align}
where $\mathcal{L}_{ent}$ means to increase the polarity inside a class, i.e., to make intra-class features converge, and $K$ means different classes. The second term $\mathcal{L}_{div}$ aims to increase the inter-class features diverge, where $g_{k}$ denotes the logits of target features.

\subsection{Proposed Framework}
Our proposed framework is shown in Figure \ref{fig:frame} and the overall objective function combines the pretrain source feature extractor with DP to protect source privacy, KD loss to mitigate BERT's overfitting issue, and IM loss to improve the UDA performance:
\begin{equation}
\begin{aligned}
 {\mathcal{L}_d} &(\theta_{E_t},\theta_{p_s}) =\mathcal{L}_{cls}(C_s(E^{dp}_s(\mathcal{X}_s)), \mathcal{Y}_s) \\ 	
&+ {\mathcal{L}_d}_{z_i^s\sim p^{dp}_s(z) }(\mathcal{Z}_s, E_t(\mathcal{X}_t)) +\mathcal{L}_{KD}(\mathcal{X}_t)+\mathcal{L}_{IM}.
	\label{DA_loss3}
\end{aligned}
\end{equation}

\textit{We adopt two backbone state-of-art UDA methods in our framework:}

(1) \textbf{Domain Adversarial Neural Network (DANN)}.
DANN imposes adversarial learning to align the source and target features. It trains a discriminator $D$ to distinguish the source and target features, while the feature extractor aims to generate domain-invariant features to fool the discriminator \cite{ganin2015unsupervised}. DANN uses a shared feature extractor $E$ on two domains with the following objective function:
\begin{equation}
\begin{aligned}
\mathcal{L}_d &= \min_{\theta_{E}} \max_{\theta_D} -\mathbb{E}^{dp}_{\mathbf{x}_i^s \sim \mathcal{D}_s}\log [D(E(\mathbf{x}_i^s))] \\
	  &- \mathbb{E}_{\mathbf{x}_i^t \sim \mathcal{D}_t}\log[1-D(E(\mathbf{x}_i^t))],
\label{eq:DA_loss}
\end{aligned}
\end{equation}

\textbf{Source-Privacy protected DA with DANN.} 
DANN uses both the source and target data to learn the shared feature extractor $E$. We sample source features from the DP source feature distribution model ($p^{dp}_s(z)$) and then the encoder $E$ for DANN is optimized by
\begin{equation}
\begin{aligned}
\mathcal{L}_d &= \min_{\theta_{E}} \max_{\theta_D} 
-\mathbb{E}_{\mathbf{z}_i^s\sim p^{dp}_s(z)}\log [D(\mathbf{z}_i^s))] \\
	  &- \mathbb{E}_{\mathbf{x}_i^t \sim \mathcal{D}_t}\log[1-D(E(\mathbf{x}_i^t))],
\label{SF_DA_loss}
\end{aligned}
\end{equation}

(2) \textbf{Conditional Adversarial Domain Adaptation (CDAN)}. 
CDAN adopts conditional adversarial learning to align the joint distribution of features and prediction of classifiers \cite{long2018conditional}, and it also has a shared feature extractor $E$ on two domains. The CDAN loss function is:
\begin{equation}
\begin{aligned}
	\mathcal{L}_d  &= \min_{\theta_{E}} \max_{\theta_D} -
	\mathbb{E}_{\mathbf{x}_i^s \sim \mathcal{D}_s}\log [D(E(\mathbf{x}_i^s), C(E(\mathbf{x}_i^s))] \\
	&- \mathbb{E}_{\mathbf{x}_i^t \sim \mathcal{D}_t}\log[1-D(E(\mathbf{x}_i^t), C(E(\mathbf{x}_i^t)))].
\end{aligned}
\end{equation}

\textbf{Source-Privacy Protected DA with CDAN.}
For the privacy version of CDAN, we also use separate encoders for source and target domains, and sample source features from the DP source feature distribution $p^{dp}_s(z)$. Then the objective function becomes this:
\begin{equation}
\begin{aligned}
	\mathcal{L}_d &= \min_{\theta_{E_t}} \max_{\theta_D} -\mathbb{E}_{\mathbf{z}_i^s \sim p^{dp}_s(z)}\log [D(\mathbf{z}_i^s, C_s(\mathbf{z}_i^s)] \\
	&- \mathbb{E}_{_i^t \sim \mathcal{D}_t}\log[1-D(E_t(\mathbf{x}_i^t), C(E_t(\mathbf{x}_i^t)))].
\end{aligned}
\end{equation}

\begin{table*}[t]
\small
\centering
\begin{tabular}{c|c|c c c c|c c c c|c c c c}
\hline
& & \multicolumn{4}{c|}{Precision} & \multicolumn{4}{c|}{Recall} & \multicolumn{4}{c}{F1}\\
\hline
& Disease & \rotatebox{90}{No adapt} & \rotatebox{90}{w source} & \rotatebox{90}{w/o source(Ours)} & \rotatebox{90}{w/o source+DP(Ours)} & \rotatebox{90}{No adapt} & \rotatebox{90}{w source} & \rotatebox{90}{w/o source(Ours)} & \rotatebox{90}{w/o source+DP(Ours)} & \rotatebox{90}{No adapt} & \rotatebox{90}{w source} & \rotatebox{90}{w/o source(Ours)} & \rotatebox{90}{w/o source+DP(Ours)} \\
\hline
& Cardiomegaly & 0.870 & 0.853 & \textbf{0.916} & 0.913 & 0.870 & 0.853 & \textbf{0.916} & 0.913 & 0.870 & 0.853 & \textbf{0.916} & 0.913\\
& Edema & \textbf{0.987} & 0.982 & 0.981 & 0.975 & \textbf{0.987} & 0.982 & 0.981 & 0.975 & \textbf{0.987} & 0.982 & 0.981 & 0.975\\
& Pneumothorax & 0.994 & 0.991 & \textbf{0.995} & 0.993 & 0.994 & 0.991 & \textbf{0.995} & 0.993 & 0.994 & 0.991 & \textbf{0.995} & 0.993\\
DANN & No finding & 0.806 & 0.721 & 0.840 & \textbf{0.849} & 0.806 & 0.721 & 0.840 & \textbf{0.849} & 0.806 & 0.721 & 0.840 & \textbf{0.854}\\
& Consolidation & 0.988 & 0.984 & \textbf{0.993} & 0.992 & 0.988 & 0.984 & \textbf{0.993} & 0.992 & 0.988 & 0.984 & \textbf{0.993} & 0.992\\
& Pneumonia & 0.984 & \textbf{0.985} & 0.960 & 0.962 & 0.984 & \textbf{0.985} & 0.960 & 0.962 & 0.984 & \textbf{0.985} & 0.960 & 0.962\\
& Fracture & 0.960 & 0.958 & 0.965 & \textbf{0.966} & 0.960 & 0.958 & 0.965 & \textbf{0.966} & 0.960 & 0.958 & 0.965 & \textbf{0.966}\\
& Pleural effusion & 0.948 & 0.948 & 0.951 & \textbf{0.952} & 0.948 & 0.948 & 0.951 & \textbf{0.952} & 0.948 & 0.948 & 0.951 & \textbf{0.952}\\
& Atelectasis & 0.855 & 0.855 & 0.914 & \textbf{0.931} & 0.855 & 0.855 & 0.914 & \textbf{0.931} & 0.855 & 0.855 & 0.914 & \textbf{0.931}\\
\hline
& Average & 0.932 & 0.919 & 0.946 & \textbf{0.948} & 0.932 & 0.919 & 0.946 & \textbf{0.948} & 0.932 & 0.919 & 0.946 & \textbf{0.948}\\
\hline
& Cardiomegaly & 0.870 & 0.864 & \textbf{0.895} & \textbf{0.895} & 0.870 & 0.864 & \textbf{0.987} & 0.895 & 0.870 & 0.864 & \textbf{0.987} & 0.895\\
& Edema & 0.987 & \textbf{0.989} & 0.987 & 0.987 & 0.987 & \textbf{0.989} & 0.987 & 0.987 & 0.987 & \textbf{0.989} & 0.987 & 0.987\\
& Pneumothorax & 0.994 & \textbf{0.997} & 0.996 & 0.996 & 0.994 & \textbf{0.997} & 0.996 & 0.996 & 0.994 & \textbf{0.997} & 0.996 & 0.996\\
CDAN & No finding & 0.806 & 0.817 & \textbf{0.884} & \textbf{0.884} & 0.806 & 0.817 & \textbf{0.884} & \textbf{0.884} & 0.806 & 0.817 & \textbf{0.884} & \textbf{0.884} \\
& Consolidation & 0.988 & 0.989 & \textbf{0.991} & \textbf{0.991} & 0.988 & 0.989 & \textbf{0.991} & \textbf{0.991} & 0.988 & 0.989 & \textbf{0.991} & \textbf{0.991}\\
& Pneumonia & 0.984 & \textbf{0.986} & 0.977 & 0.977 & 0.984 & \textbf{0.986} & 0.977 & 0.977 & 0.984 & \textbf{0.986} & 0.977 & 0.977\\
& Fracture & 0.960 & 0.965 & \textbf{0.971} & \textbf{0.971} & 0.960 & 0.965 & \textbf{0.971} & \textbf{0.971} & 0.960 & 0.965 & \textbf{0.971} & \textbf{0.971}\\
& Pleural effusion & 0.948 & 0.971 & \textbf{0.974} & \textbf{0.974} & 0.948 & 0.971 & \textbf{0.974} & \textbf{0.974} & 0.948 & 0.971 & \textbf{0.974} & \textbf{0.974}\\
& Atelectasis & 0.855 & 0.887 & \textbf{0.902} & \textbf{0.902} & 0.855 & 0.887 & \textbf{0.902} & \textbf{0.902} & 0.855 & 0.887 & \textbf{0.902} & \textbf{0.902}\\
\hline
& Average & 0.932& 0.940 & \textbf{0.953} & \textbf{0.953} & 0.932& 0.940 & \textbf{0.963} & 0.953& 0.932& 0.940 & \textbf{0.963}& 0.953\\
\hline
\end{tabular}
\caption{Precision, recall, and F1 score of DANN and CDAN under the following conditions: "no adapt" means directly evaluate the source model on target data; "w source" means applying the adaptation method with source data and evaluating the target model on target data; "w/o source" denotes our proposed method excluding DP-pretrain; "w/o source+DP" is the full version of our proposed method.}
\label{tab:med}
\end{table*}

\section{Experiments}
In this section, we implement our proposed source-privacy preserving UDA method and perform various experiments and visualization to evaluate the effectiveness of our proposed method on two datasets of the medical report label extraction task. We first verify the main task performance of UDA through precision, recall, and F1-score. Then we compare the performance under different privacy losses.

\subsection{Datasets and Hyperparameters}
For the pulmonary disease label extraction problem, we select two related datasets: MIMIC-CXR Database (MIMIC) \cite{johnson2019mimic} and Open-i collection from Indiana University taken from \url{openi.nlm.nih.gov} (Open-i) \cite{demner2010combining}. We manually select the common disease labels to mitigate the label difference between two datasets and filter the corresponding medical report findings. The common labels we select are \textit{cardiomegaly, edema, pneumothorax, no finding, consolidation, pneumonia, fracture, pleural effusion, and atelectasis}. Therefore, the main task is to infer correct findings from the clinical report. We select the findings section in the medical report as the input text in both datasets. Note that each medical report may have multiple positive findings. Therefore this task can be viewed as a multi-label classification problem. For the MIMIC dataset, there are three types of labels for negative (0), positive (1), and uncertain (-1). We merge the uncertain labels into the negative class for simplicity purposes. For the Open-i dataset, we extract the labels from the Mesh section. Since the Open-i dataset (1858 samples) is much smaller than the MIMIC dataset (115708 samples) after our filtration, we only apply the adaptation in one direction, i.e., from MIMIC to Open-i.
\par
We opt for precision, recall, and F1-score as the main task performance metrics to evaluate the performance of UDA on the pulmonary disease label extraction problem, as it can be abstracted as a multi-label classification task.
\par
To apply adversarial learning and distinguish the features from source and target domains, the discriminator is implemented as a 3-layer fully connected network. The AdamW optimizer \cite{loshchilov2017decoupled} is adopted to update all models (source and target feature extractors, classifier, and discriminator). The learning rate for pretraining the source feature extractor and classifier is $5\times10^{-5}$. For adapting the target feature extractor and discriminator, the learning rate is set to $1\times10^{-5}$.

\subsection{Pulmonary Disease Label Extraction}
Since samples from both datasets (MIMIC and Open-i) may have multiple labels, we cannot simply use the accuracy as the evaluation metric. Instead, we report the results in precision, recall, and F1-score. We mainly compare the performance of UDA on three settings: with source (adapting with source data), without-source (adapting with GMM resampled features), and without-source DP (adapting with DP-GMM resampled features) in this task. We report the evaluation metrics on each common class between two datasets. Table \ref{tab:med} summaries the results of pulmonary disease multi-label classification. Note that many results appear to have the same value is due to the rounding issue. We only evaluate the adaptation in one direction: from MIMIC to Open-i because the MIMIC dataset has much more samples than the Open-i dataset (115708 compared to 1858), and there are nine classes in total. For each adaptation method, we compare four scenarios: "no adapt" means we directly evaluate the source model on the target data as the baseline; "with source" denotes adaptation with source data; "without source" is our proposed method except the DP pretraining; and "without source+DP" is the complete version of our proposed method. From Table \ref{tab:med}, we can observe that our proposed method performs better than the "no adapt" and "with source" version in most cases, especially in the "cardiomegaly" and "no finding" classes. In other classes, our proposed method evens with the rest of the classes in that the baseline results are already saturated and have little space for improvements, e.g., edema, pneumothorax, consolidation, pneumonia, and fracture. In our implementation, the "without-source DP" version's privacy budget is set to 1. The "without-source DP" version is slightly inferior to the without-source version in some classes. This performance is as expected because the DP version introduces some noise in pretraining and leads the GMM to deviate from the source feature distribution.
\par
Interestingly, our proposed method constantly performs better than the with-source version in this pulmonary disease classification task. We hypothesis that the MIMIC dataset is very noisy with many inaccurate labels as pointed out by \cite{hao2020inaccurate}. Hence, the source feature distributions might be highly irregular and have many outliers. The GMM clustering of source features in our proposed method helps to regularize the source feature distributions. Thus, the resampled features are clustering better than the source features and this phenomenon can assist the target data adaptation. 
\par
\begin{figure}[t]
\includegraphics[width=.43\textwidth]{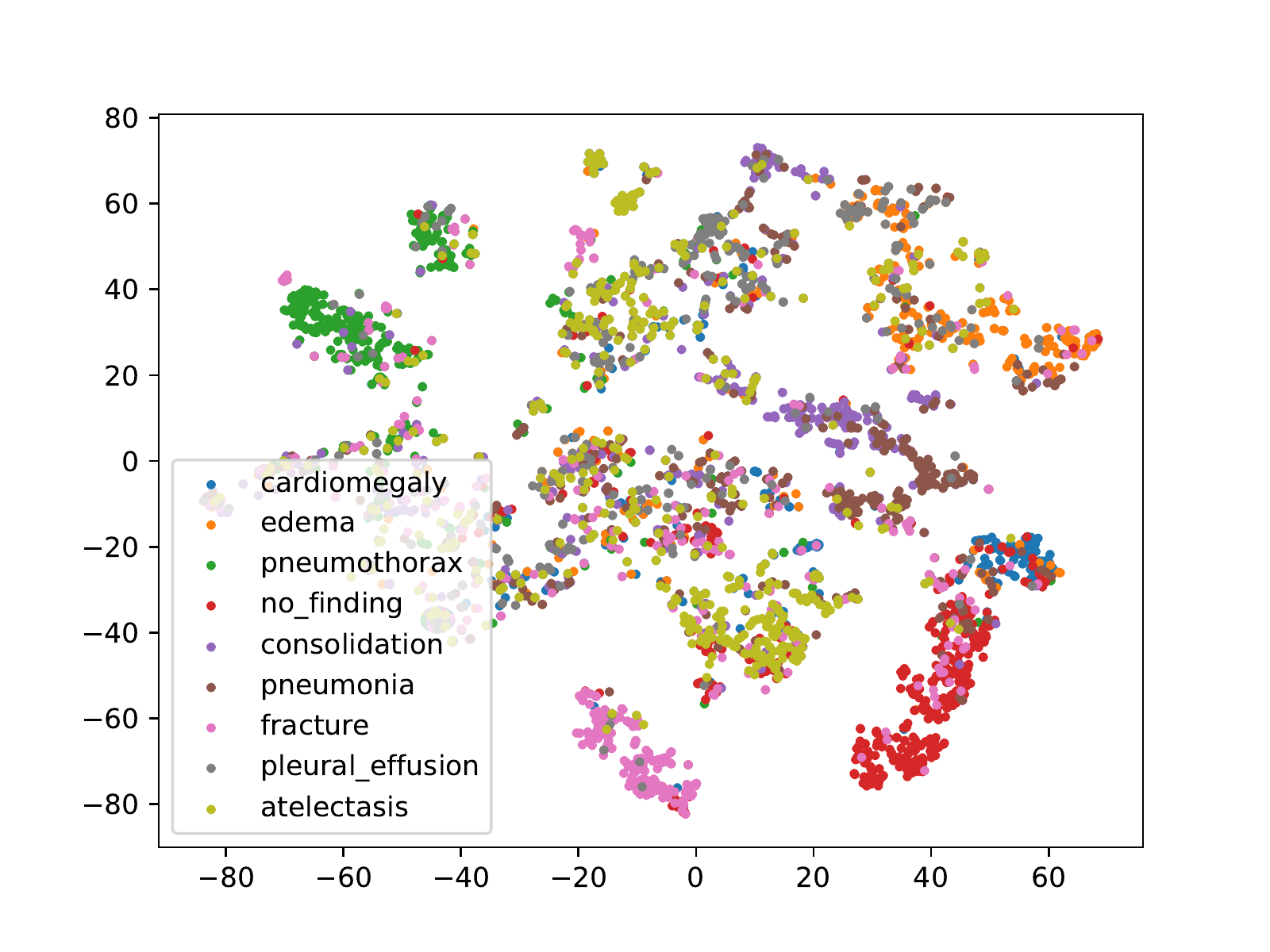}
\caption{TSNE visualization of target features before adaptation.}
\label{fig:before_gmm}
\vfill
\includegraphics[width=.43\textwidth]{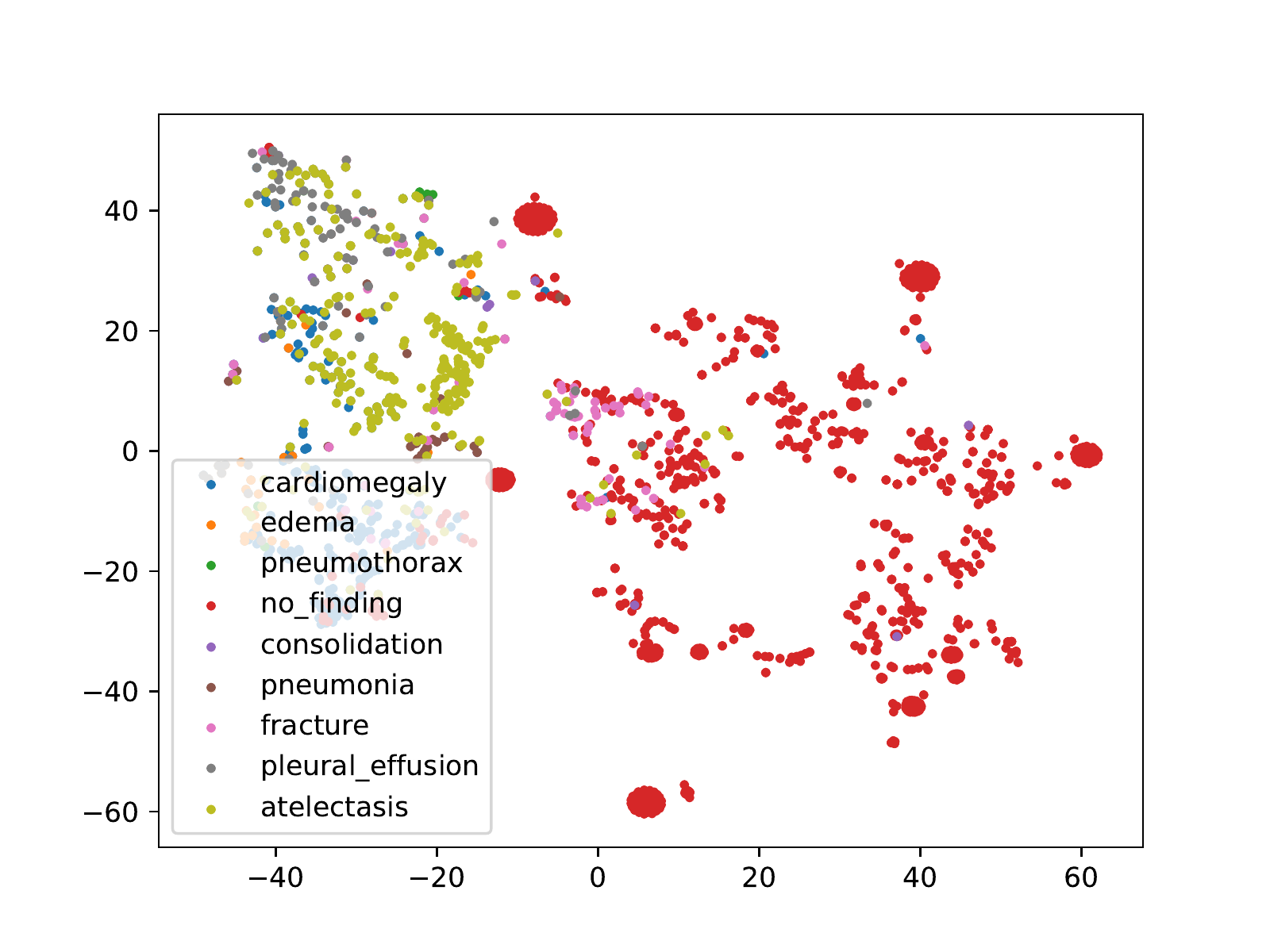}
\caption{TSNE visualization of target features after with source CDAN+KD adaptation.}
\label{fig:openi_w_src}
\vfill
\includegraphics[width=.43\textwidth]{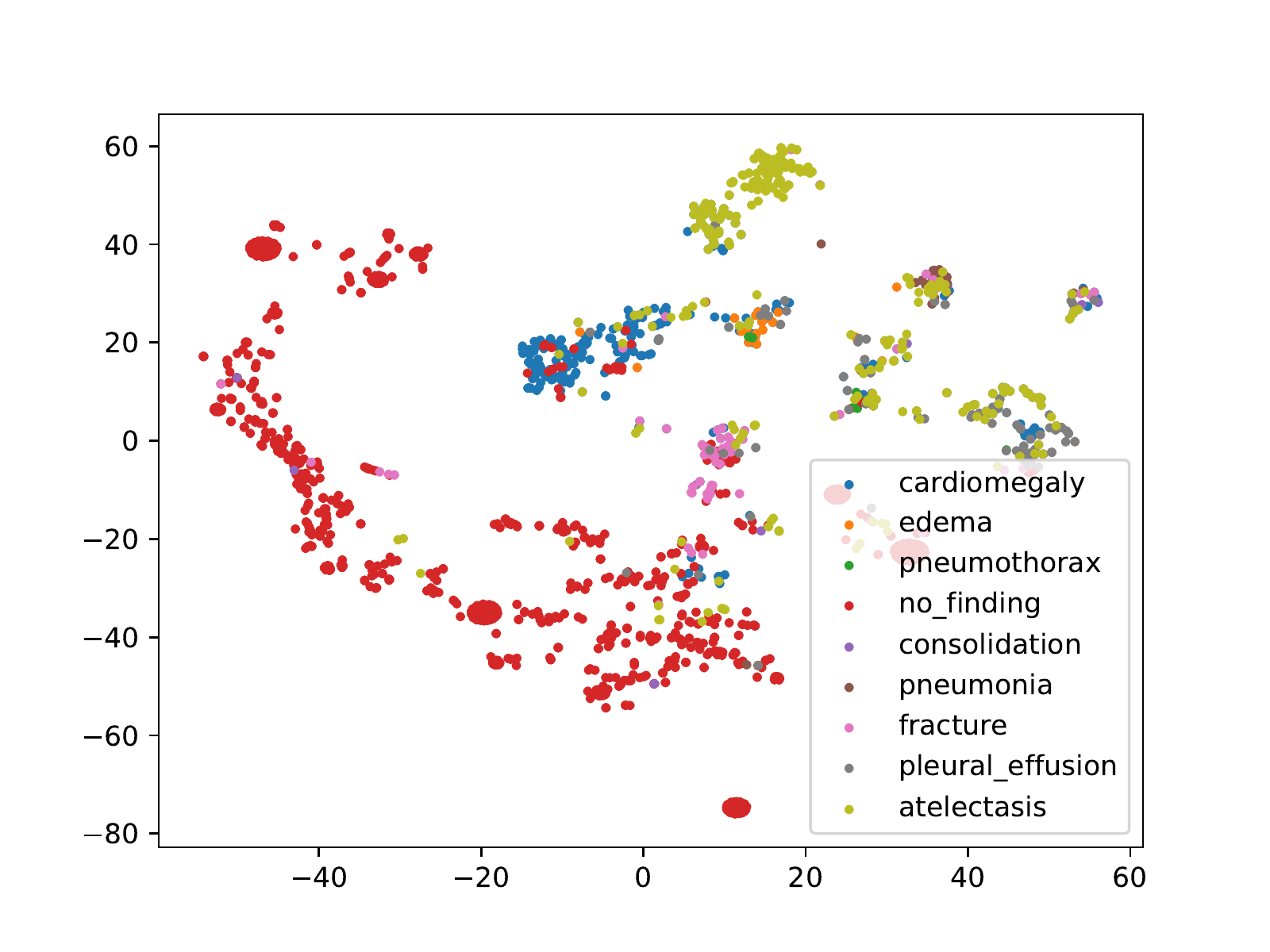}
\caption{TSNE visualization of target features after without source CDAN+KD+DP adaptation.}
\label{fig:openi_wo_src}
\end{figure}

To verify our hypothesis, we visualize the target features in 3 scenarios: 1) the target encoder before adaptation; 2) the target encoder after CDAN adaptation with source data; 3) the target encoder after CDAN+DP adaptation without source data (out proposed method); as shown in Figure \ref{fig:before_gmm}, \ref{fig:openi_w_src}, and \ref{fig:openi_wo_src} respectively. We can observe that features within a class converge better than before, and features among different classes become more divergent. This phenomenon confirms the effectiveness of GMM in our proposed method.

\begin{table}[t]
\centering
\begin{tabular}{c|c|c c c c c}
\hline
& Disease & \multicolumn{5}{c}{F1}\\
\hline
& $\epsilon$ & \rotatebox{0}{2.5} & \rotatebox{0}{5} & \rotatebox{0}{7.5} & \rotatebox{0}{15} & non-DP \\
\hline
 & Cardiomegaly & 0.908 & 0.912 & 0.910 & 0.916 & \textbf{0.987} \\ 
 & Edema & \textbf{0.987} & \textbf{0.987} & 0.986 & 0.986 & \textbf{0.987} \\ 
 & Pneumothorax & 0.994 & 0.993 & 0.994 & 0.994 & \textbf{0.996} \\ 
 & No finding & 0.874 & 0.876 & 0.869 & 0.876 & \textbf{0.884} \\ 
CDAN & Consolidation & 0.988 & \textbf{0.991} & 0.990 & \textbf{0.991} & \textbf{0.991} \\ 
 & Pneumonia & 0.979 & \textbf{0.980} & \textbf{0.980} & \textbf{0.980} & 0.977 \\ 
 & Fracture & 0.969 & \textbf{0.971} & 0.969 & 0.969 & \textbf{0.971} \\ 
 & Pleural effusion & 0.965 & 0.967 & 0.966 & 0.967 & \textbf{0.974} \\ 
 & Atelectasis & 0.899 & 0.899 & 0.899 & 0.899 & \textbf{0.902} \\ 
\hline
& Average & 0.951 & 0.952 & 0.951 & 0.953 & \textbf{0.963}\\
\hline
\end{tabular}
\caption{F1 scores of CDAN under different privacy loss $\epsilon$. Note that the smaller the $\epsilon$, the stronger the privacy is protected.}
\label{tab:privacy_loss}
\end{table}

\begin{figure}[h]
    \centering
    \includegraphics[width=.5\textwidth]{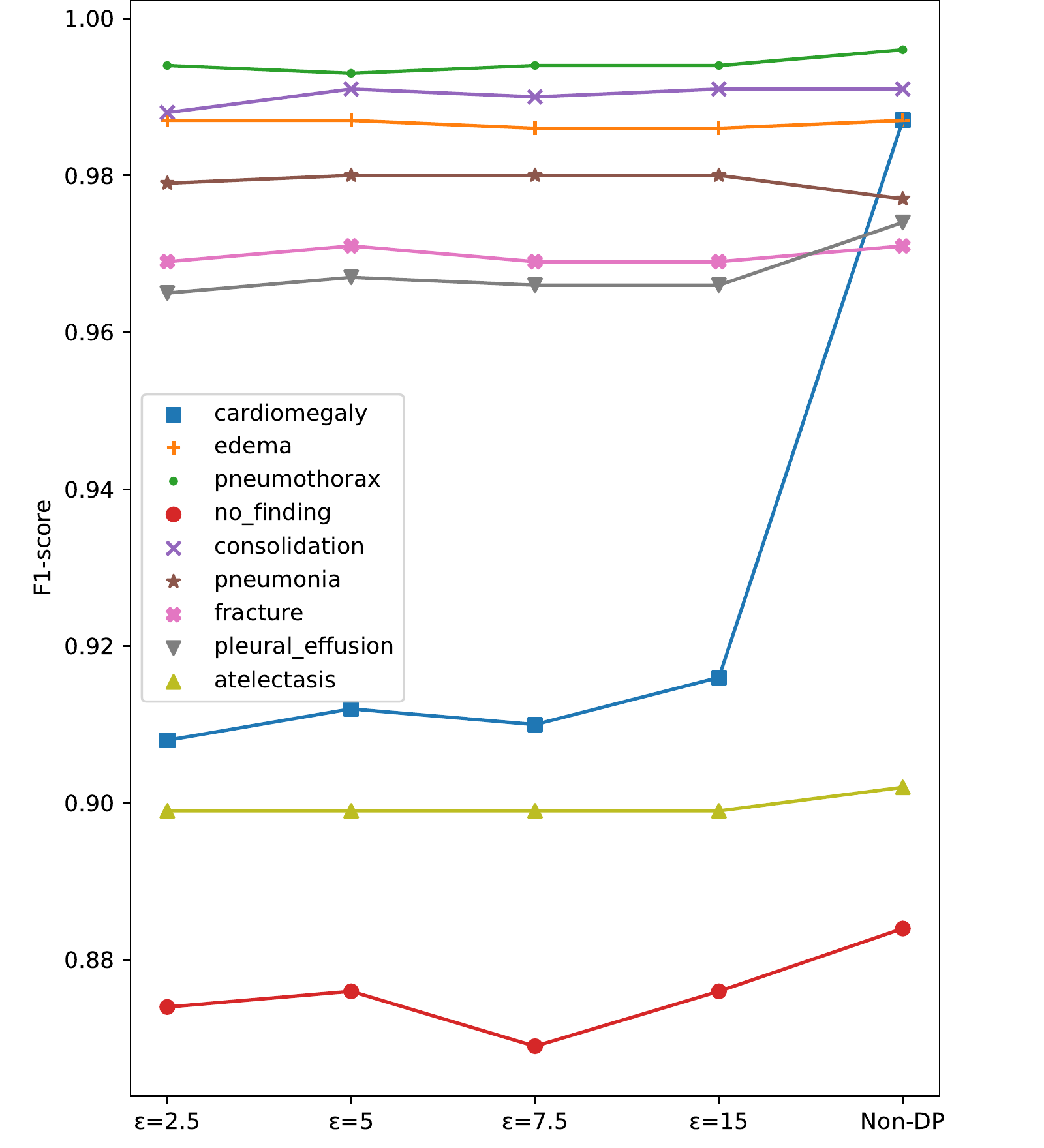}
    \caption{$\epsilon$ vs F1. $\epsilon$ is the privacy loss.}
    \label{fig:eps_vs_perf}
\end{figure}

\subsection{Performance under Different Privacy Loss}
In the DP-SGD method, the privacy budget, denoted as $\epsilon$, is to quantify how much we tolerate the output of a model on one database $d$ to be different from the output of the same model on a second database $d'$ that only has one different training sample. The smaller the $\epsilon$ is, the stronger the privacy will be \cite{papernot2018scalable}. We evaluate the main task performance of pulmonary disease classification under different privacy budgets. The privacy loss is measured by the moments accountant (MA) from \cite{abadi2016deep}. In our implementation, we adopt the Opacus\footnote{https://opacus.ai/} library to implement the DP-pretrain and privacy loss accountant. We only perturb the last transformer layer in the BERT encoder to perform the DP-pretrain due to the memory limit on our server. We compare the F1-score of each disease with different privacy loss as the F1-score combines the advantage of precision and recall. The non-DP version denotes that we only resample the GMM model of source feature distribution to adapt the target model, which is equivalent to the "without source" version in Table \ref{tab:med}. We adopt four different values (2.5, 5, 7.5, 15) of $\epsilon$ to evaluate adaptation performance. Here, we only report the F1 score since it combines the benefits of precision and recall and better represents the overall performance. In our implementation, the privacy loss is accumulated from the pretraining process through MA. The results are summarized in Table \ref{tab:privacy_loss} and Figure \ref{fig:eps_vs_perf}. From Table \ref{tab:privacy_loss} and Figure \ref{fig:eps_vs_perf}, we can observe that the DP-pretraining does not deteriorate the performance too much, with only some small changes in each class. Applying the DP-pretraining, for classes cardiomegaly, no finding and pleural effusion, the DP-pretrain versions become worse. Other classes are almost the same. This is consistent with our expectations.

\begin{figure}[h]
\centering
\includegraphics[width=.43\textwidth]{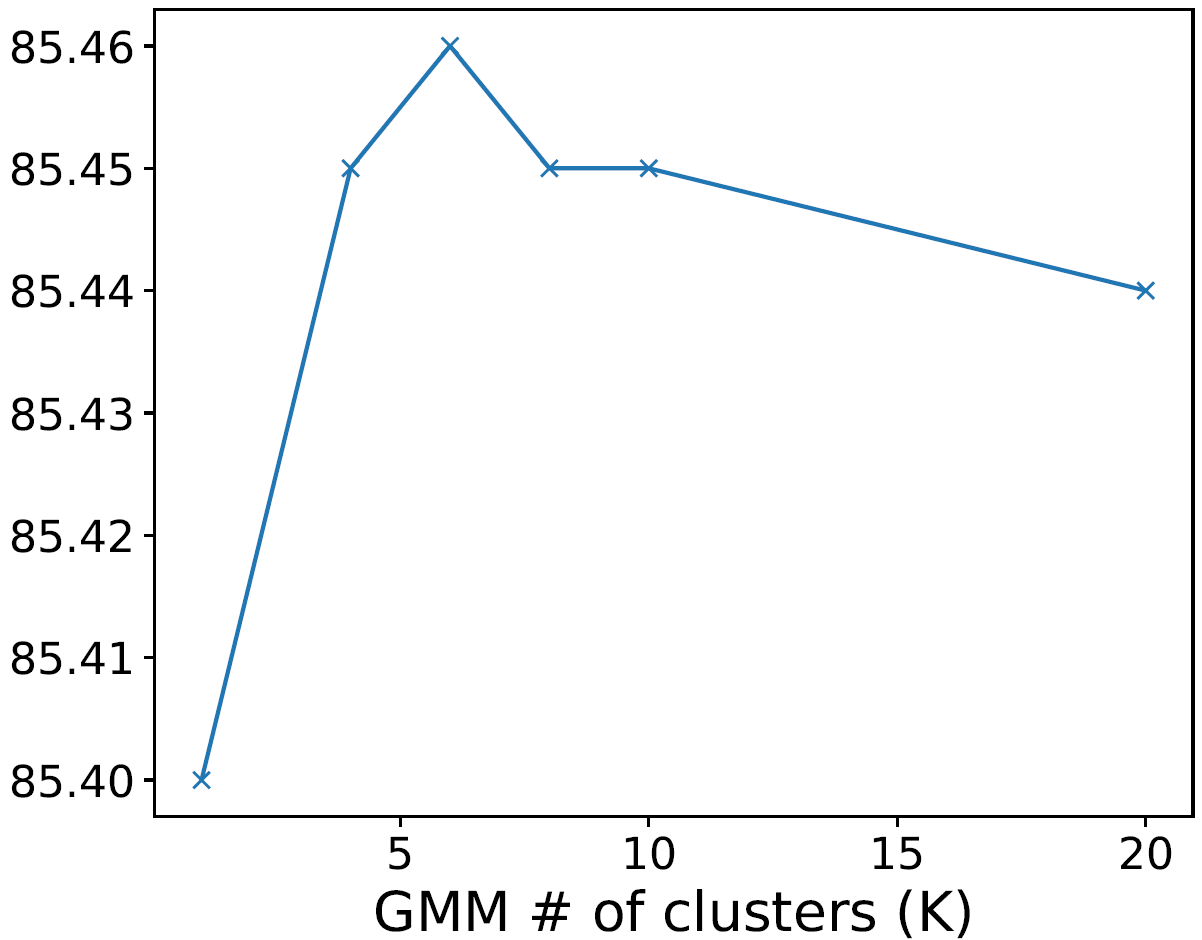}
\caption{Average adaptation F1-score (\%) vs. number of clusters (K) on DANN+DP (class no finding).}
\label{fig:gmm_k}
\end{figure}

\subsection{Discussion of Cluster Number (K) in GMM}
Another factor that can influence the performance of our proposed model is the number of clusters K in the GMM. From our experiments based on the without source version of DANN, the factor K does impact the adaptation performance (class no finding) as shown in Figure \ref{fig:gmm_k}. The highest F1-score 0.8546 appears at K = 6, and the performance deteriorates with larger K values indicating that the number of GMM clusters is not the larger, the better. The best results appear around 5 or 6. We also adopt the same K value in other clinical text-dataset adaptations.

\section{Conclusion}
In this paper, we mainly study how to preserve the source data privacy in the unsupervised domain adaptation on clinical text data. We discuss how the privacy leaks when sharing the source data to target clients for adaptation and when confined to only sharing the source feature distribution. To defend against this source privacy leakage, we propose a source privacy-preserving unsupervised domain adaptation framework that adopts differential privacy in pretraining the source model to reduce the risk of source-data privacy leakage. Our proposed method is compatible with many existing unsupervised domain adaptation methods. We implement our framework on two clinical text datasets to perform the pulmonary disease multi-label classification task. Our experiment results have verified that our proposed method can protect the source data privacy with little sacrifice on the utility.

\bibliographystyle{IEEEtranS.bst}
\bibliography{dpuda.bib}

\end{document}